\documentclass[conference, a4paper]{IEEEtran}
\IEEEoverridecommandlockouts
\usepackage{geometry}
\usepackage{cite}
\usepackage{amsmath,amssymb,amsfonts}
\usepackage{algorithmic}
\usepackage{graphicx}
\usepackage{textcomp}
\usepackage{wrapfig}
\usepackage{picinpar}
\usepackage{url}
\usepackage{flushend}
\usepackage[latin1]{inputenc}
\usepackage{colortbl}
\usepackage{soul}
\usepackage{multirow}
\usepackage{pifont}
\usepackage{color}
\usepackage{alltt}
\usepackage[hidelinks]{hyperref}
\usepackage{enumerate}
\usepackage{siunitx}
\usepackage{breakurl}
\usepackage{epstopdf}
\usepackage{pbox}
\usepackage{subfig}
\usepackage[linesnumbered,ruled,vlined]{algorithm2e}
\usepackage{booktabs}
\usepackage[table,xcdraw]{xcolor}

\begin{document}
\title{DDCNN: A Promising Tool for Simulation-to-reality UAV Fault Diagnosis
\thanks{\textcopyright{} 20XX IEEE.  Personal use of this material is permitted. Permission from IEEE must be obtained for all other uses, including reprinting/republishing this material for advertising or promotional purposes, collecting new collected works for resale or redistribution to servers or lists, or reuse of any copyrighted component of this work in other works.}
\thanks{Wei Zhang, Shanze Wang and Xiaoyu Shen are with the College of Information Science and Technology, Eastern Institute of Technology, Ningbo, China.
        {(e-mail: zhw@eitech.edu.cn, shanze.wang@u.nus.edu,
        xyshen@eitech.edu.cn)}
    }
\thanks{Junjie Tong and Yunfeng Zhang are with the Department of Mechanical Engineering, National University of Singapore, Singapore 117575, Singapore.
	{(e-mail: e0240715@u.nus.edu, mpezyf@nus.edu.sg)}
	}%
\thanks{Fang Liao is with Temasek Laboratories, National University of Singapore, Singapore 117411, Singapore.
	{ (e-mail: tsllf@nus.edu.sg)}
	}%
}

\author{Wei Zhang, Shanze Wang, Junjie Tong, Fang Liao, Yunfeng Zhang, Xiaoyu Shen
}

\maketitle

\begin{abstract}

Identifying the fault in propellers is important to keep quadrotors operating safely and efficiently. The simulation-to-reality (sim-to-real) UAV fault diagnosis methods provide a cost-effective and safe approach to detecting propeller faults. However, due to the gap between simulation and reality, classifiers trained with simulated data usually underperform in real flights. In this work, a novel difference-based deep convolutional neural network (DDCNN) model is presented to address the above issue. It uses the difference features extracted by deep convolutional neural networks to reduce the sim-to-real gap. Moreover, a new domain adaptation (DA) method is presented to further bring the distribution of the real-flight data closer to that of the simulation data. The experimental results demonstrate that the DDCNN+DA model can increase the accuracy from 52.9\% to 99.1\% in real-world UAV fault detection.

\end{abstract}


\IEEEpeerreviewmaketitle

\section{Introduction}

\begin{figure}[htp]
	\centering
	\includegraphics[width=0.8\linewidth]{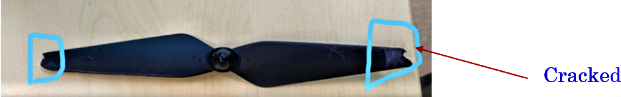}
	\caption{Example of broken propeller.}
	\label{propeller}
\end{figure}

\IEEEPARstart{Q}{uadrotor} unmanned aerial vehicles (UAVs), become increasingly popular in various fields, such as search-and-rescue and homeland security \cite{shraim2018survey}. However, their propellers may be vulnerable to damage during operation  (see Fig. \ref{propeller}) due to unforeseen events, such as collisions with buildings or trees. Such damage can impede the UAV's capability to accomplish its mission and may even lead to significant damage to the UAV itself and to people in the surrounding area. Therefore, it is essential to keep monitoring the conditions of the quadrotor propellers. Identifying propeller problems at an early stage allows for a timely recall of the UAVs, and hence reducing further damages.

Based on the use of system models or not, UAV fault diagnosis methods classified as either model-based or model-free approaches. The model-based approach \cite{guzman2019actuator}, \cite{Ma2019}, \cite{Gai2022} requires first building a model of the system and then comparing the model output with the actual output to identify propeller failures. Therefore, an accurate system model of a real UAV is the key to successful fault diagnosis. However, due to noise and installation errors, building an accurate system is not trivial and requires a great deal of expert knowledge and extensive experiments. 

Compared to model-based approaches, model-free methods do not rely on a system model. Among model-free approaches, data-driven approaches \cite{iannace2019fault}, \cite{park2022multiclass} have become increasingly popular in recent years. These approaches utilize flying data and fault labels to train a fault classifier. The fault classifier is built on deep neural networks (DNNs) \cite{hinton2006reducing}, which can extract features from the sensor input and perform classification simultaneously. Such an end-to-end mechanism bypasses the model-building process and, therefore, does not require much knowledge of UAV system.

In contrast to other fields of data-driven fault diagnosis, such as bearing fault diagnosis \cite{Zhang2018}, the collection of UAV faults is difficult because flying with a broken propeller can result in a drone crash. To reduce the risk of real-world data collection, it is much safer to collect training data in a simulation and then deploy the well-trained classifier to a real UAV. This approach is also known as simulation-to-reality (sim-to-real), which is a common approach for learning data-driven controllers for robots \cite{Jemin2019}, \cite{zhang2022ipaprec}. However, due to noise and mounting errors, a simulated UAV cannot have the same behavior as a real UAV under the same commands, which poses a challenge for a fault classifier trained with simulated data \cite{tong2023machine}.

In this research, we focus on developing a data-driven classifier that can bridge the sim-to-real gap and achieve high fault-diagnosis accuracy in real flight. To reach this goal, we propose the difference-based convolutional neural network (DDCNN) that learns difference features from the tested sample and all-healthy category. Although the difference features are learned from simulated data, they can generalize well in real flight. Moreover, a simple but effective domain adaptation (DA) method is presented to align the distribution of the all-healthy samples from simulation and real flight. It can further reduce the sim-to-real gap and improve the classification performance. To sum up, the contributions of this study are outlined as follows,
\begin{itemize}
	\item The novel DDCNN model is proposed to improve the sim-to-real performance of the UAV fault classifier. The idea of DDCNN, which exploits the difference features between faulty and healthy samples, offers a promising direction for the field of fault diagnosis. 
    \item A new DA method is presented to help DDCNN model further reduce the sim-to-real gap.
    \item Using DDCNN and DA, the simulated data trained propeller fault classifier can increase the accuracy from 52.9\% to 99.1\% in real-flight fault diagnosis.
\end{itemize}

The rest of this work is structured as follows. A concise overview of relevant literature is given in Section \ref{RW}, then the problem description is presented in Section \ref{Preliminaries}. In Section \ref{approach}, we introduce the proposed DDCNN model and DA approaches, followed by a comprehensive performance analysis through model training and evaluation in Section \ref{Implementation}. Lastly, the conclusions of this work are drawn in Section \ref{Conclusion}.

\section{Related works}\label{RW}
\subsection{Data-driven UAV fault diagnosis}
Data-driven UAV fault diagnosis approaches train DNNs as classifiers to learn fault diagnosis skill from the manually-labelled flying data. The flying data usually contains the state information, such as roll, pitch, yaw, angular velocities and accelerations, and the rotational speeds of propellers. A new model named improved one-dimensional deep residual shrinkage network with a wide convolutional layer (1D-WIDRSN) was presented for UAV fault diagnosis \cite{yang2021intelligent}. Trained with real flight data, the classifier can perform accurate fault diagnosis on the test data collected from the same flights. Similarly, Park et al. \cite{park2022multiclass} proposed the stacked pruning sparse denoising auto-encoder to detect UAV fault. Due to the use of denoising auto-encoder, the classifier performed well in the noisy scenario. Li et al. \cite{Li2022} utilized siamese neural network \cite{Taigman2014} to diagnose the fixed-wing UAV when only few training samples were avaliable. However, this method only showed advantage in limited training samples. When the size of the training samples increased, its performance was similar to that of the conventional classifier such as support vector machine \cite{Hearst1998}.

\subsection{Sim-to-real transfer of DNNs}
Although sim-to-real transfer is new in the field of UAV fault diagnosis, it has gained much attention from other fields, especially in robot control \cite{Tai2017}. Zhang et al. \cite{zhang2021learn} trained a mobile robot using LiDAR as a distance sensor in a simulated scenario, and the trained DNN controller can control the robot in the real world with different LiDAR configurations. For quadrotors, Rubi et al. \cite{Rubi2021} trained a quadrotor to learn path-following and obstacle-avoidance tasks in in simulation, and the trained model can be used by a real quadrotor. However, for the above approaches, the gap between simulation and reality is small due to the use of simulated high-fidelity LiDAR sensor. Therefore, the data-driven controller learned from simulated scenarios can control the real robot without any fine-tuning.

For sim-to-real robot control based on vision, the robot's observations in simulation may differ from those in the real world \cite{zielinski20213d}. To reduce the gap between simulation and reality, the domain randomization method for sim-to-real drone race task was presented \cite{Loquercio2020}. However, for our UAV fault diagnosis task, the sim-to-real gap is mainly caused by differences in the UAV body, i.e., mounting errors and noise, rather than by observations received by the on-board sensors.

\section{Preliminaries}\label{Preliminaries}

\subsection{Problem description}
The quadrotor propeller fault diagnosis problem can be model as a classification problem. In this problem, there are five categories, including one all-healthy category and four fault categories, which are listed in Table \ref{labels}. The input to the classifier is made up of the flying data of the quadrotor. The same as \cite{zhang2023}, in this paper, at step time $t$, the input $x_t$ comprise the angular accelerations and the square of the rotational speed of the propellers as follows,
\begin{equation}\label{input_eq}
\begin{aligned}
x_t=\left[\begin{matrix}{\dot{p}}_{t-T}&{\dot{p}}_{t-T+1}&\cdots&{\dot{p}}_t\\{\dot{q}}_{t-T}&{\dot{q}}_{t-T+1}&\cdots&{\dot{q}}_t\\{\dot{r}}_{t-T}&{\dot{r}}_{t-T+1}&\cdots&{\dot{r}}_t\\\omega_{1, t-T}^2&\omega_{1, t-T+1}^2&\cdots&\omega_{1, t}^2\\\omega_{2, t-T}^2&\omega_{2, t-T+1}^2&\cdots&\omega_{2, t}^2\\\omega_{3, t-T}^2&\omega_{3, t-T+1}^2&\cdots&\omega_{3, t}^2\\\omega_{4, t-T}^2&\omega_{4, t-T+1}^2&\cdots&\omega_{4, t}^2\\\end{matrix}\right].
\end{aligned}
\end{equation}
where  $(\dot{p},\dot{q},\dot{r})$  denotes quadrotor angular accelerations; $\omega_{i=1,2,3,4}$ denote the rotational speeds of the four propellers. $T$ is time-window length. It determines how much historical information is taken into account in the input. The purpose of the classifier is to label the input sample based on their output as one of the five categories shown in Table \ref{labels}.

\begin{table}[htp]
\caption{\label{labels} Labels of five fault categories.}
\centering
\begin{tabular}{llllll}
\hline
\hline
Propeller No. & label 1 & label 2 & label 3 & label 4 & label 5\\ \hline
Propeller 1 & healthy & faulty & healthy & healthy & healthy \\
Propeller 2 & healthy & healthy & faulty & healthy & healthy\\
Propeller 3 & healthy & healthy & healthy & faulty & healthy \\
Propeller 4 & healthy & healthy & healthy & healthy & faulty \\
\hline
\hline
\end{tabular}
\end{table}

In this work, the classifier can only be trained with samples generated in simulation and all-healthy samples from real flight because collecting samples with the broken propeller from real flight is dangerous. Once trained, the classifier will be employed to detect fault for real flight. The sim-to-real classification problem can be viewed as cross-domain classification problem. The simulation domain is referred to as source domain, while the real-world domain is regarded as target domain. This cross-domain classification problem is challenging because of the domain gap induced by noise and installation error. The objective of this paper is to train a classifier with simulated data to achieve high accuracy on real flight data.

\section{APPROACH}\label{approach}
\begin{figure}[htp]
	\centering
	\includegraphics[width=0.9\linewidth]{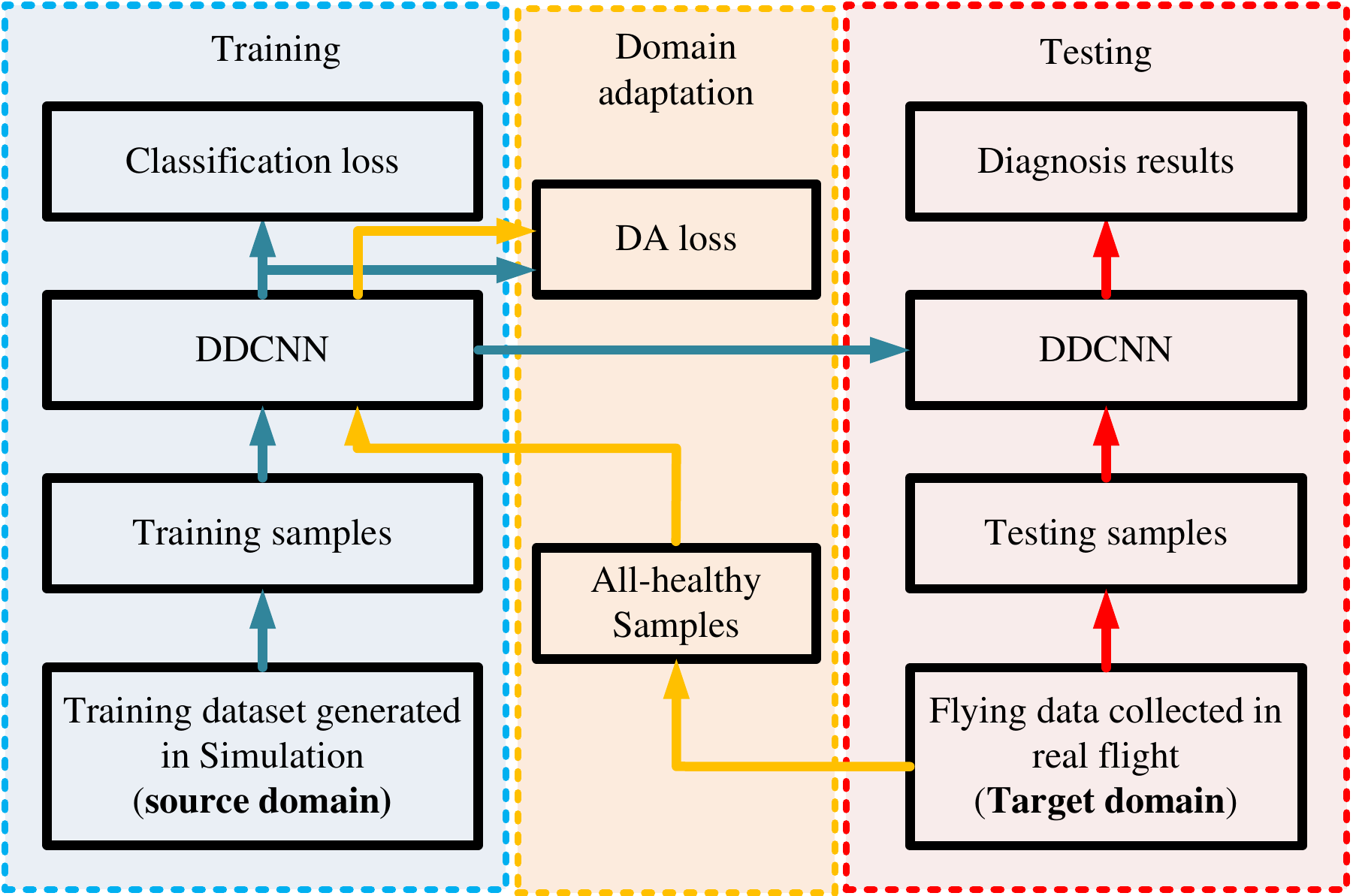}
	\caption{Framework of the proposed approach.}
	\label{Framework}
\end{figure}

In this section, a new method for monitoring the health condition of quadrotor propellers is presented. The overall framework of the proposed approach is given in Fig. \ref{Framework}. As shown, DDCNN is proposed as the fault classifier. The training data comprises simulated flight data and all-healthy samples collected from an individual flight. To enhance the generalization capability of the DDCNN model in actual flight, a new domain adaptation method is introduced. Once trained, the DDCNN model can be employed to monitor the status of the UAV in real-world situations. 

\subsection{DDCNN model}\label{DDNN}
\begin{figure*}[htpb]
    \centering
	  \subfloat[]{
       \includegraphics[width=0.25\linewidth]{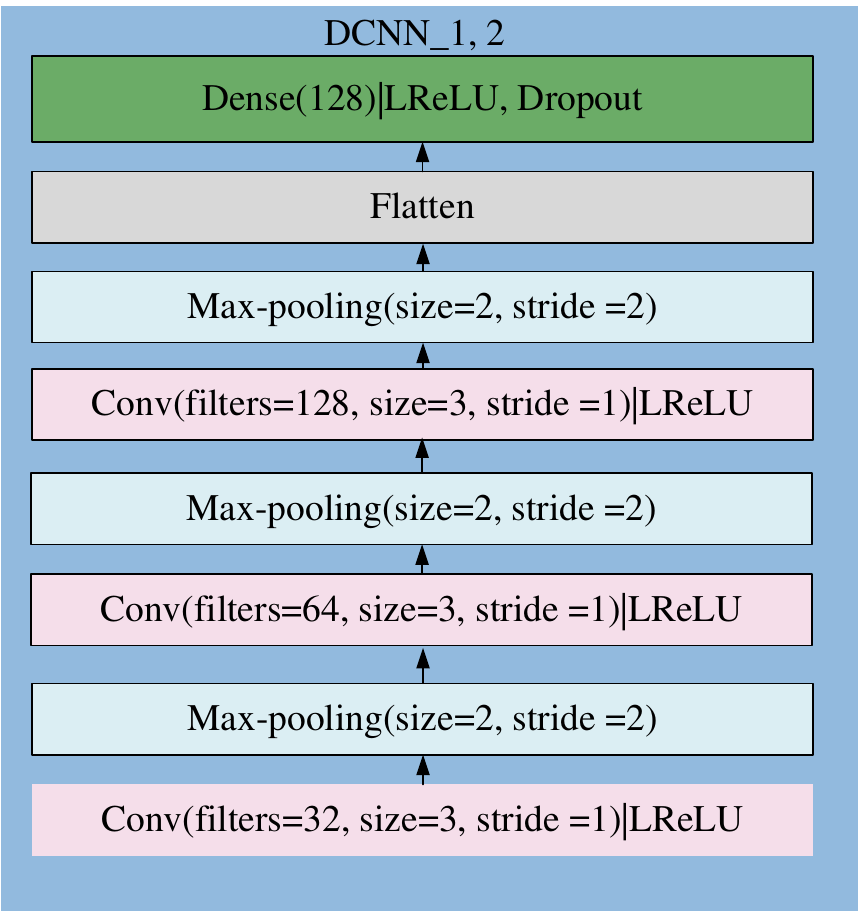}\label{DCNN12}}
	  \subfloat[]{
        \includegraphics[width=0.36\linewidth]{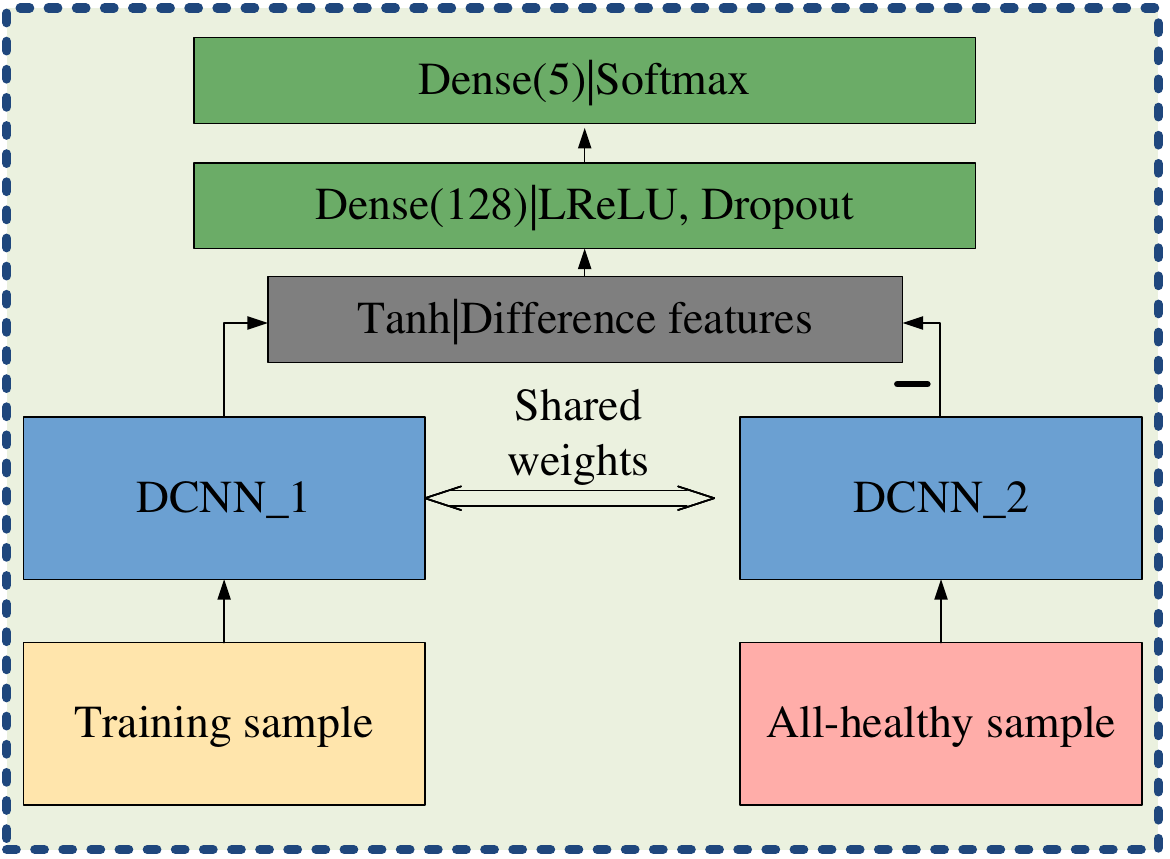}\label{train_DDCNN}}
        \subfloat[]{
        \includegraphics[width=0.36\linewidth]{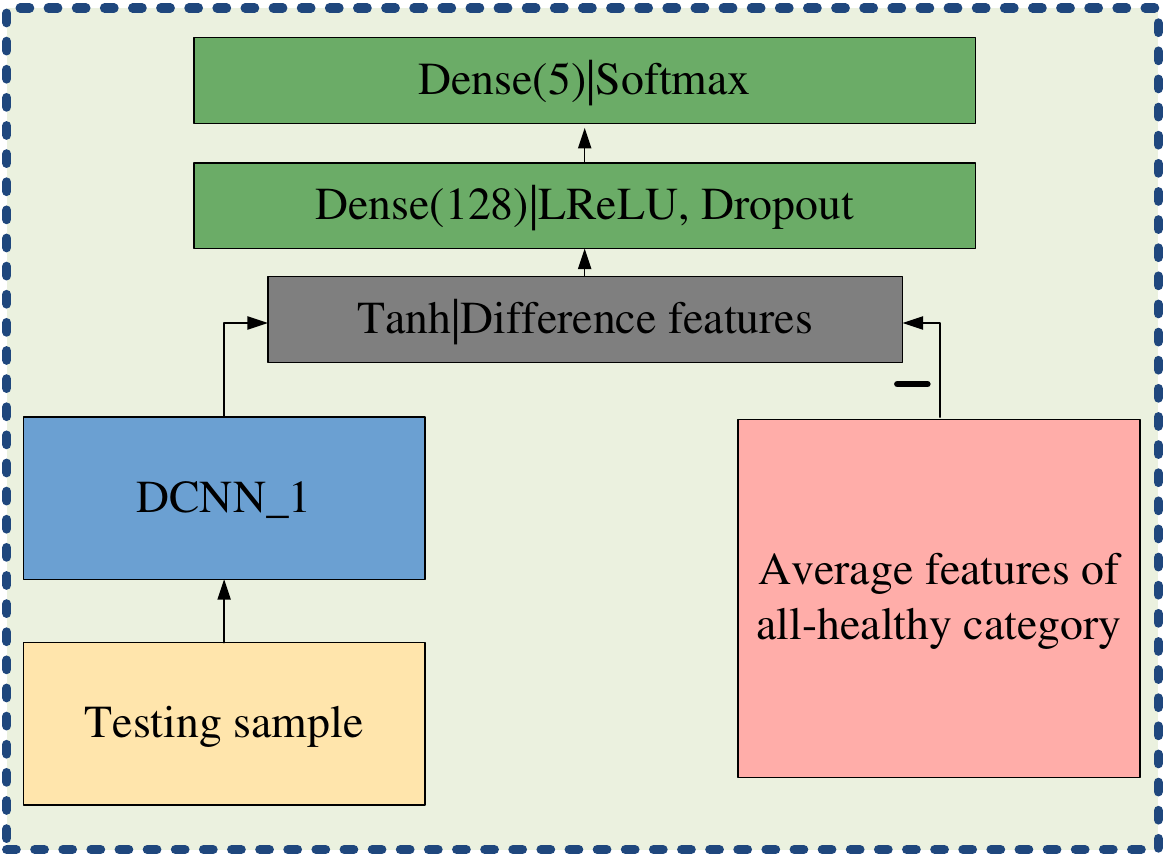}\label{test_DDCNN}}
	\caption{Network struture of DDCNN. (a) The network structure of DCNN\_1 and DCNN\_2. (b) The network structure of DDCNN used for training. (c) The network structure of DDCNN used for testing.}
	\label{DDCNN}
\end{figure*}

The architecture of the DDCNN model is depicted in Fig. \ref{DDCNN}. As shown, similar to the structure of the Siamese neural network (SNN) \cite{Li2022}, it contains two identical deep convolutional neural networks (DCNNs), namely DCNN\_1 and DCNN\_2 , parameterized by $\phi_1$, $\phi_2$, respectively. In other words, DCNN\_1 and DCNN\_2 share the same network structure and weights, i.e., $\phi_1=\phi_2$. It should be noted that the only similarity between DDCNN and SNN is the use of the twin structure. the overall network structure, loss function, training and testing procedures of DDCNN are completely different from those of SNN.

As shown in Fig. \ref{DCNN12}, the DCNN models contain three convolutional layers. After each convolutional layer, a max-pooling operation is performed. The features learned by the convolutional layers are flattened and fed into a fully-connected layer. The 1D vector outputted by the fully-connected layer is treated as the features of the input sample. Due to domain change, the features of the same-category samples from different domain, i.e., simulation and real flight, are different. However, the differences between the faulty features and the all-healthy features can be similar in different domain. Hence, our DDCNN model assumes that the differences between faulty features and all-healthy features are similar in different domain. Under this  assumption, the features used for fault classification are the differences between the features of the diagnosed sample and the features of the all-healthy category.

 The calculations of the difference features $\phi_d$ vary in the training and testing phases. During training, as shown in Fig. \ref{train_DDCNN}, the input to DCNN\_1, i.e., $x_i$, is the training samples sampled from any fault categories, while the input to DCNN\_2, i.e., $x_j$, is the training samples sampled from the all-healthy category. After forward propagation in each DCNN, the difference features $\phi_{d,train}$ used in training are calculated as follows,
\begin{equation}
    \phi_{d,train} = \text{tanh}(\phi_1(x_i)-\phi_2(x_j))
\end{equation}
where $\phi_1(x_i)$ denotes the features outputted by the DCNN\_1 on sample $x_i$; $\phi_2(x_j)$ represents the features outputted by the DCNN\_2 on sample $x_j$. 
$\text{tanh}(\cdot)$ is the tanh activation function, which controls the range of each difference feature be $(-1,1)$. 

With the difference features $\phi_{d}$, as shown in Fig. \ref{train_DDCNN} and Fig. \ref{test_DDCNN}, a classifier with a dense hidden layer is used to compute the probability of each type of fault. The output layer adopts the Softmax function to calculate the discrete probability distributions of the five health conditions of the quadrotor. The formula for the Softmax function is:
\begin{equation}\label{softmax}
\begin{aligned}
q\left(z_j\right)=\text{softmax}\left(z_j\right)=\frac{e^{z_j}}{\sum_{i=1}^{5}e^{z_i}}.
\end{aligned}
\end{equation}
where $z_j$ represents the logits of the $j$-th neuron in the output layer.

The classification loss for training the DDCNN model is the cross-entropy between the approximated softmax output distribution and the corresponding target class distribution. More formally, denoting the target distribution and the estimated distribution outputted by DDCNN as $p(x)$ and $q(x)$ respectively, the cross-entropy loss $\mathcal{L}_{c
}$ between $p(x)$ and $q(x)$ is expressed as follows:
\begin{equation}\label{cross entropy}
\begin{aligned}
\mathcal{L}_{c
}= H(p(x),q(x))=-\sum_{x}p(x)\log q(x).
\end{aligned}
\end{equation}
\subsection{Domain adaptation based regularization}\label{DA}
When using fault classifiers for fault diagnosis, their performance may be affected by the different distribution of data in the target and source domains. To address this problem, domain adaptation (DA) approaches are frequently employed to align the distribution of the target domain data with that of the source domain data. However, typical DA approaches require access to the full range of unlabeled data from the target domain \cite{Liu2022}, which is not feasible in our case due to the risks associated with flying a quadrotor with a broken propeller. Therefore, our DA method aims to utilize only the all-healthy samples from the target domain to extract transferable features, which allows for a safer approach. Similar to \cite{Lu2017}, only the all-healthy samples from the target domain are utilized for DA. Our DA method assumes the cross-domain variation way for samples of different faults is similar. Under this assumption, when the distributions of all-healthy source and target domain data are brought closer, the distance between the distributions of the source and target domain for other categories is also reduced. In this paper, the difference features in DDCNN serve as the features to perform domain adaptation. Due to the use of the difference features, the target distribution of the features in both domains are zero-centered. Therefore, different from \cite{Lu2017}, we do not need to calculate the domain distance, such as the MMD (maximum mean discrepancy) distance, and minimize the domain distance. In this paper, we only need to make each element of the difference features zero, and the DA loss used in this paper is shown as follows,
\begin{equation}\label{DA loss}
\begin{aligned}
\mathcal{L}_{DA}=\frac{1}{n_{AH}}\sum_{x_j\in\mathcal{X}_{AH}}\lVert\phi_{d,train}\left(x_j\right)\rVert_2^2.
\end{aligned}
\end{equation}
where $\mathcal{X}_{AH}$ denotes the group of the all-healthy real-flight samples, which can be collected from an individual flight. $n_{AH}$ is the number of samples in $\mathcal{X}_{AH}$. Minimizing the DA loss can make the difference features of all-healthy samples in target domain closer to zero and their distribution closer to that of all-healthy samples in source domain. By integrating the classification loss and DA loss together, the model loss function is as follows,
\begin{equation}\label{loss}
\begin{aligned}
\mathcal{L} = \mathcal{L}_{c}+\lambda\mathcal{L}_{DA}
\end{aligned}
\end{equation}
where the factor $\lambda$ is a constant that weighs the contribution of the DA loss.
\subsection{Difference features in testing}\label{DDCNN_test}
During testing, as shown in Fig. \ref{train_DDCNN}, the input $x_i$ to DCNN\_1 is the sample to be tested (this sample can be from any fault categories), while DCNN\_2 and its input are replaced by the pre-computed feature representation $\bar{\phi_2}$. $\bar{\phi_2}$ represents the features of the all-healthy category extracted by DCNN\_2. With $\bar{\phi_2}$, the difference features calculated during the testing are as follows,
\begin{equation}\label{d_test}
    \phi_{d,test} = \text{tanh}(\phi_1(x_i)-\bar{\phi_2})
\end{equation}
In this paper, $\bar{\phi_2}$ is estimated using the average features of all healthy samples as follows,
\begin{equation}\label{phi_average}
    \bar{\phi_2} =\frac{1}{n_{AH}}\sum_{x_j\in\mathcal{X}_{AH}}{\phi_2(x_j)}
\end{equation}
 With (\ref{d_test}) and (\ref{phi_average}), during testing, only the forward computation of $\phi_1(x_i)$ is performed for calculating the difference features. The training and testing procedure of our algorithm can be found in Algorithm \ref{algorithm1}.

\begin{algorithm}[t]
\SetAlgoLined
\textbf{Model Training}\;
 Initialize parameters of DDCNN $\theta$\;
 Input: training Dataset A, all-healthy real-flight Dataset B, and testing Dataset C\;
 Copy the all-healthy category in Dataset A and Dataset B into Dataset D and E\;
 \For{\upshape{epoch}$=1$, 2, \ldots, }
 { \For{\upshape{training step}$=1$, 2, \ldots, }
  { Randomly sample $m$ samples from Dataset A, B, D and E into a mini-batch $\mathcal{M}_A$, $\mathcal{M}_B$, $\mathcal{M}_D$ and $\mathcal{M}_E$ \;
  Calculating the cross entropy loss using $\mathcal{M}_A$ and $\mathcal{M}_D$ with (\ref{cross entropy})\;
  Calculating the DA loss using $\mathcal{M}_B$ and $\mathcal{M}_E$ with (\ref{DA loss})\;
  Updating the $\theta$ by minimizing total loss $\mathcal{L}$ using gradient decent\;
  }}
\textbf{Model Testing}\;
 Replace the DCNN\_2 with the average features of all-healthy category using (\ref{phi_average})\;
Classify the testing samples in Dataset C using (\ref{d_test}) and (\ref{softmax}).
\caption{Training and testing procedure of DDCNN with Domain Adaptation}
\label{algorithm1}
\end{algorithm}

\section{Implementation and test}\label{Implementation}
In this section, we trained the DDCNN model and assessed its sim-to-real performance using real flight data. The training data for the DDCNN model was generated using a simulation model presented in \cite{tong2023machine}. Once trained, the DDCNN model was applied to classify faults on real flight data.

\subsection{Dataset Description}
\begin{figure}[t]
	\begin{minipage}{.52\linewidth}
		\subfloat[]{\centering\includegraphics[width=0.8\linewidth]{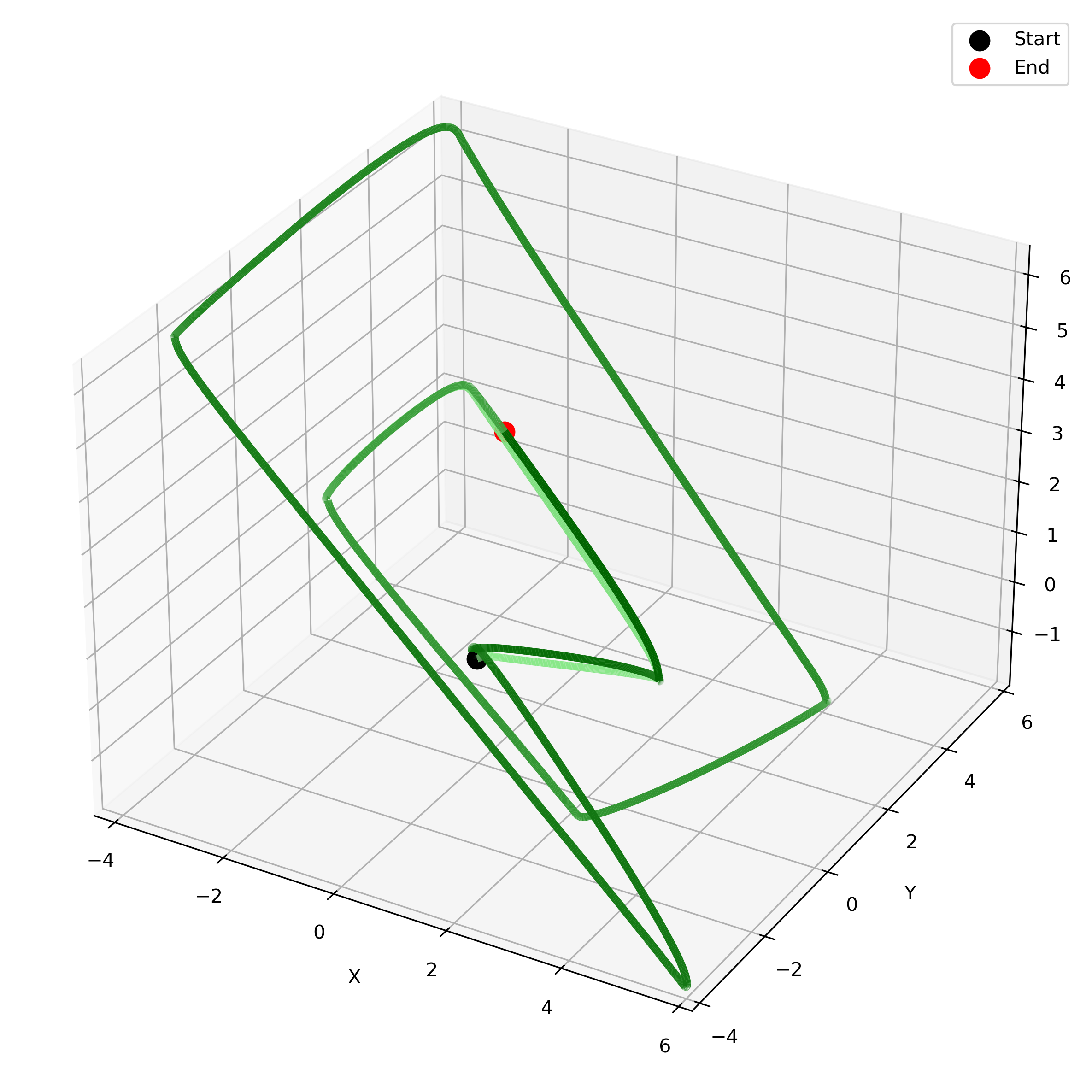}\label{sim_traj}}\\[-2.5ex]
       	\subfloat[]{\centering\includegraphics[width=0.8\linewidth]{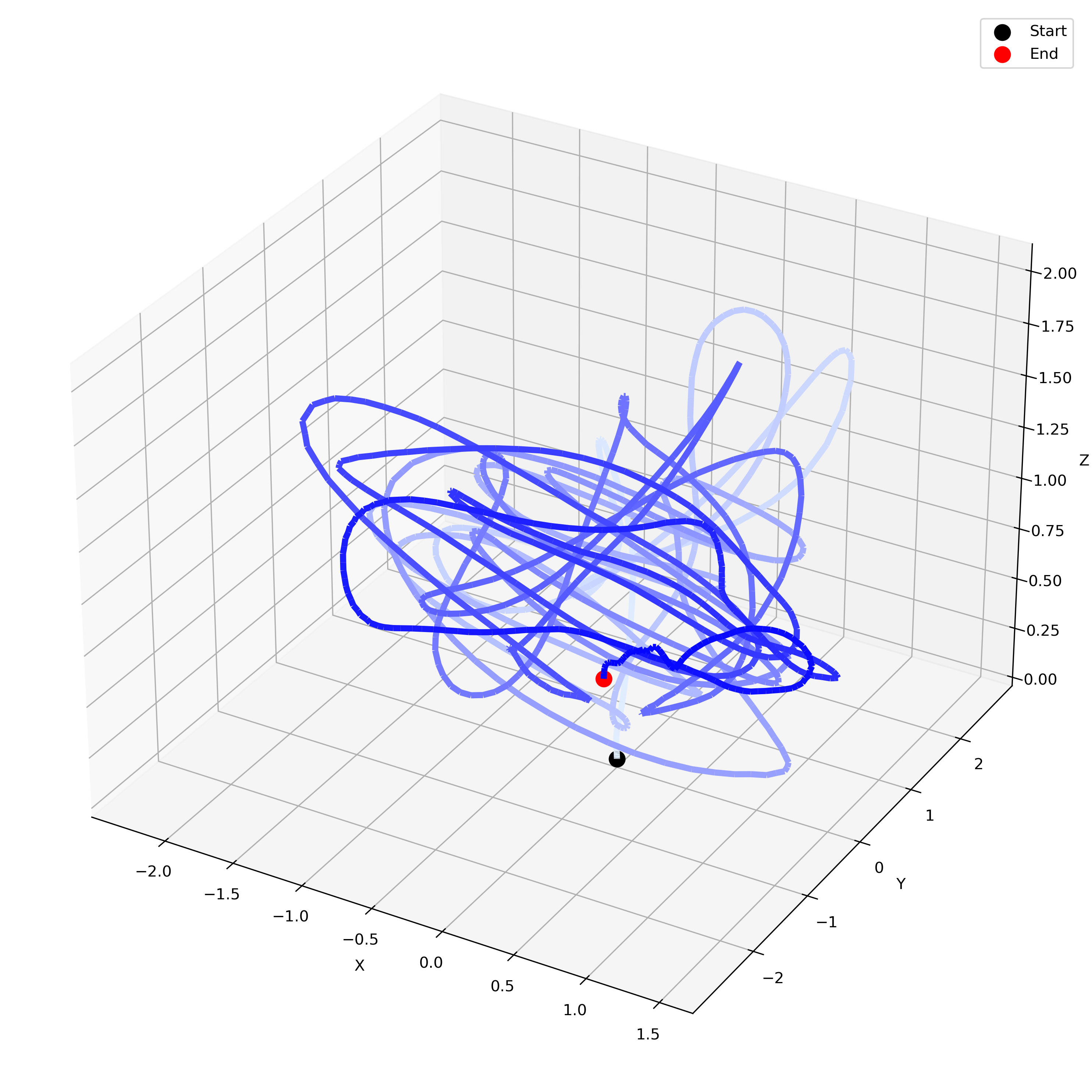}\label{real_traj}}
	\end{minipage}
    \begin{minipage}{.44\linewidth}
	\subfloat[]{\centering\includegraphics[width=0.9\linewidth]{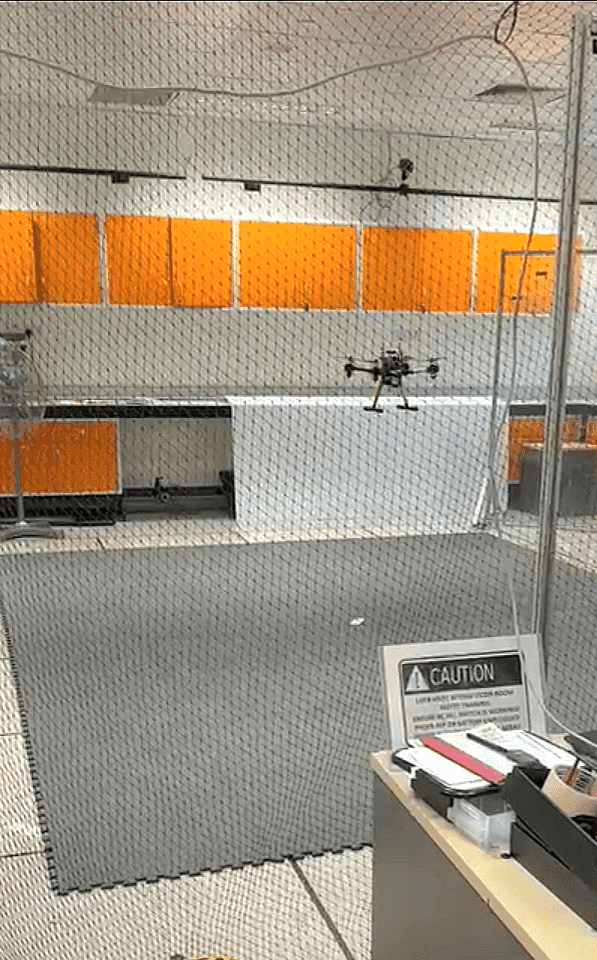}\label{real_exp}}
	\end{minipage}%
	\caption{Data collection in simulation and real flight. (a) One of the UAV trajectory for data collection in simulation. (b) One of the UAV trajectories for data collection in real flight. (c) Lab scenario for data collection in real flight. }
	\label{data_collection}
\end{figure}

As shown in Table \ref{real_dataset}, three datasets, i.e., Dataset A, B and C, are generated or collected in this paper. Dataset A was generated in simulation using the simulator introduced in \cite{tong2023machine}. This simulator employs LSTM to learn the faulty propeller model from data collected via load cells. It then integrates the learned model into a UAV simulator to generate flight data under propeller fault conditions. Dataset A covers five categories, and each category has 3000 samples. To collect samples for each fault category, the simulated UAV was required to fly to the designated waypoints using the on-board controller, and one of the UAV trajectories is shown in Fig \ref{sim_traj}. The trajectory is visualized using a color gradient that transitions from light to dark hues, corresponding to the progression from the initial to the final position. The origin and terminus of the path are demarcated by black and red markers, respectively, to enhance clarity and facilitate interpretation of the motion sequence. The waypoints were the same for the five categories.

Besides, Dataset B was utilized for training as well. As shown in Fig. \ref{real_exp}, it was collected from a real flight under the all-healthy conditions. During data collection, a human pilot randomly flew the quadrotor in an indoor scenario for around five minutes. We chose random maneuvers to assess the classifier's generalization performance across various flight conditions. Due to space constraints in the room, we conducted UAV flights within a limited area of 30$\text{m}^2$ space. However, this was not problematic as position information is not utilized for fault diagnosis.  Consequently, 3000 samples of the all-healthy category were collected. Dataset B is used for different purposes in training and testing. During training, it was used for performing domain adaptation. During testing, it was used to pre-compute the features for the real-world all-healthy category. 

\begin{figure}[htpb]
    \centering
	  \subfloat[Broken propellers]{
       \includegraphics[width=0.6\linewidth]{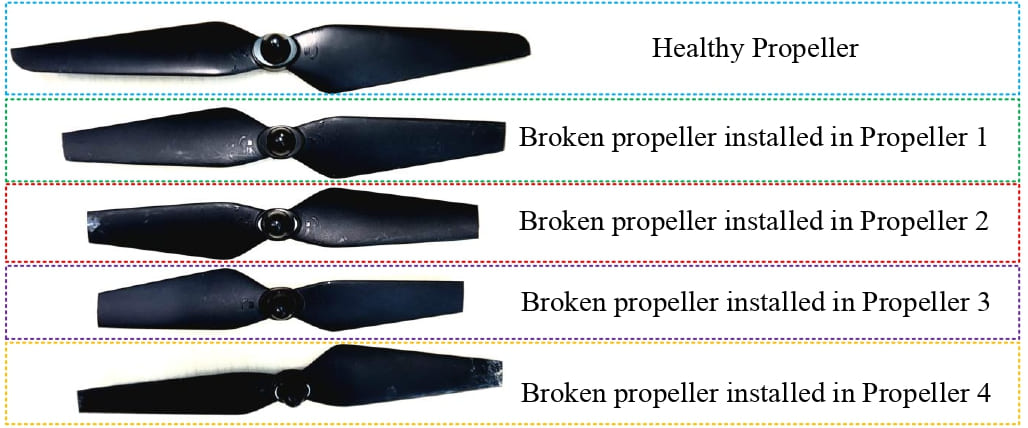}\label{bp}}
        \\
	    \subfloat[Real UAV model]{
        \includegraphics[width=0.5\linewidth]{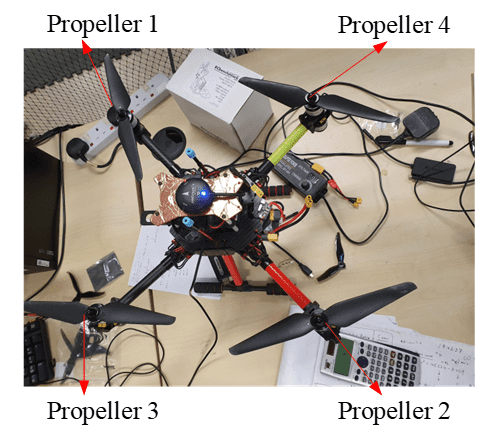}\label{real_UAV}}
	\caption{The real UAV model and broken propellers used in real-world data collection.}
	\label{real_propeller}
\end{figure}

Dataset C is the dataset used for testing. It was collected from real flight and contains five categories and has 800 samples in each category. The same as Dataset B, the UAV was controlled by a human pilot. The flying trajectories were different for each category, and each flight took about two minutes. One of the flying trajectories is shown in Fig. \ref{real_traj}. Before collecting data for each faulty category, one healthy propeller was replaced by a broken propeller in the labelled position, as shown in Fig. \ref{real_propeller}. It should be noted that the damaged levels of the broken propellers in the four positions are different.

\begin{table*}[htp]
\centering
\caption{\label{real_dataset} Description of quadrotor datasets generated for training and testing.}
\begin{tabular}{|c|c|c|ccccc|}
\hline
\multirow{3}{*}{Dataset Name} &
  \multirow{3}{*}{Data source} &
  \multirow{3}{*}{Purpose} &
  \multicolumn{5}{c|}{Category label (fault   location)} \\ \cline{4-8} 
 &
   &
   &
  \multicolumn{1}{c|}{1 (None)} &
  \multicolumn{1}{c|}{2 (Propeller 1)} &
  \multicolumn{1}{c|}{3 (Propeller 2)} &
  \multicolumn{1}{c|}{4 (Propeller 3)} &
  5 (propeller 4) \\ \cline{4-8} 
 &
   &
   &
  \multicolumn{5}{c|}{Number of samples} \\ \hline
A &
  Simulation &
  Training &
  \multicolumn{1}{c|}{3000} &
  \multicolumn{1}{c|}{3000} &
  \multicolumn{1}{c|}{3000} &
  \multicolumn{1}{c|}{3000} &
  3000 \\ \hline
B &
  Real flight &
  Training for DA &
  \multicolumn{1}{c|}{3000} &
  \multicolumn{1}{c|}{-} &
  \multicolumn{1}{c|}{-} &
  \multicolumn{1}{c|}{-} & -
   \\ \hline
C &
  Real flight &
  Testing &
  \multicolumn{1}{c|}{800} &
  \multicolumn{1}{c|}{800} &
  \multicolumn{1}{c|}{800} &
  \multicolumn{1}{c|}{800} &
  800 \\ \hline
\end{tabular}
\end{table*}
\subsection{Training details}
As the proposed model mainly utilizes DDCNN model and DA-based regularization, we refer to our model as DDCNN+DA model. Beisdes, to evaluate the effect of the proposed DA method, the DDCNN model without DA is also trained and tested. Moreover, to evaluate the effect of the difference features in DDCNN, we remove the difference operation in the DDCNN model, i.e., the DCNN\_2 branch is removed in Fig \ref{train_DDCNN}, and refer to this model as the DCNN model. The DCNN model with DA was not implemented because the presented DA method relies on the difference features of the DDCNN model, which is not suitable for the DCNN model.

The hyperparameters of training are listed in Table \ref{training}. The GPU used for training is RTX 2080Ti, which took about two minutes for each training process. To evaluate the repeatability of each model, the training procedure was repeated ten times with different random seeds.

\begin{table}[htp]
\caption{\label{training} Hyperparameter settings.}
\centering
\renewcommand{\arraystretch}{1.5}
{\begin{tabular}{ll}
\hline
\hline
Hyperparameters           & Value  \\ \hline
Mini-batch   size         & 128    \\
Learning   rate           & $5\times10^{-4}$ \\
Maximum   training epochs & 10     \\
Optimizer                 & Adam   \\
Dropout   rate            & 20\%  \\
Factor of DA loss: $\lambda$            & $0.2$  \\
\hline
\hline
\end{tabular}}
\end{table}

\subsection{Results and comparison study}
Once trained, the DDCNN+DA model can be employed for fault classification tasks. We conducted a comparative analysis of DDCNN+DA against several models: DCNN, 1D-WDRSN \cite{yang2021intelligent}, and DDCNN models.  For each model, the training procedure is repeated ten times, and the average test accuracy was used for comparison. Table \ref{real_acc} summarizes the testing accuracy of ten runs for each model on the real-flight Dataset C. The DCNN and 1D-WDRSN models achieved average accuracies of 52.9\% and 59.4\%, respectively. This difference highlights a significant performance gap when using simulated versus real-flight data. However, the average accuracy improves substantially to 85.5\% with the DDCNN model. This indicates the difference features learned from simulation data are similar to their real-flight counterparts. Moreover, incorporating DA with the DDCNN model further increases the average accuracy to 99.1\%. This promising result demonstrates that our proposed DA method can effectively narrow the simulation-to-reality gap.

\begin{table}[htp]
\renewcommand{\arraystretch}{1.5} 
\centering
\caption{\label{real_acc}Accuracy of the DCNN,1D-WDRSN, DDCNN and DDCNN+DA on  testing Dataset C (real flight).}
\begin{tabular}{lllll}
\hline
\hline
Method & DCNN & 1D-WIDRSN & DDCNN & DDCNN+DA \\ \hline
\multicolumn{1}{c}{Accuracy} & \multicolumn{1}{c}{52.9$\pm$9.0\%} & \multicolumn{1}{c}{59.4$\pm$0.9\%} & \multicolumn{1}{c}{85.5$\pm$4.5\%} & \multicolumn{1}{c}{99.1$\pm$0.5\%}\\
\hline
\hline
\end{tabular}
\end{table}

\subsection{Visualization of the difference features}\label{vis}
\begin{figure}[htp]
	\centering
	\includegraphics[width=0.8\linewidth]{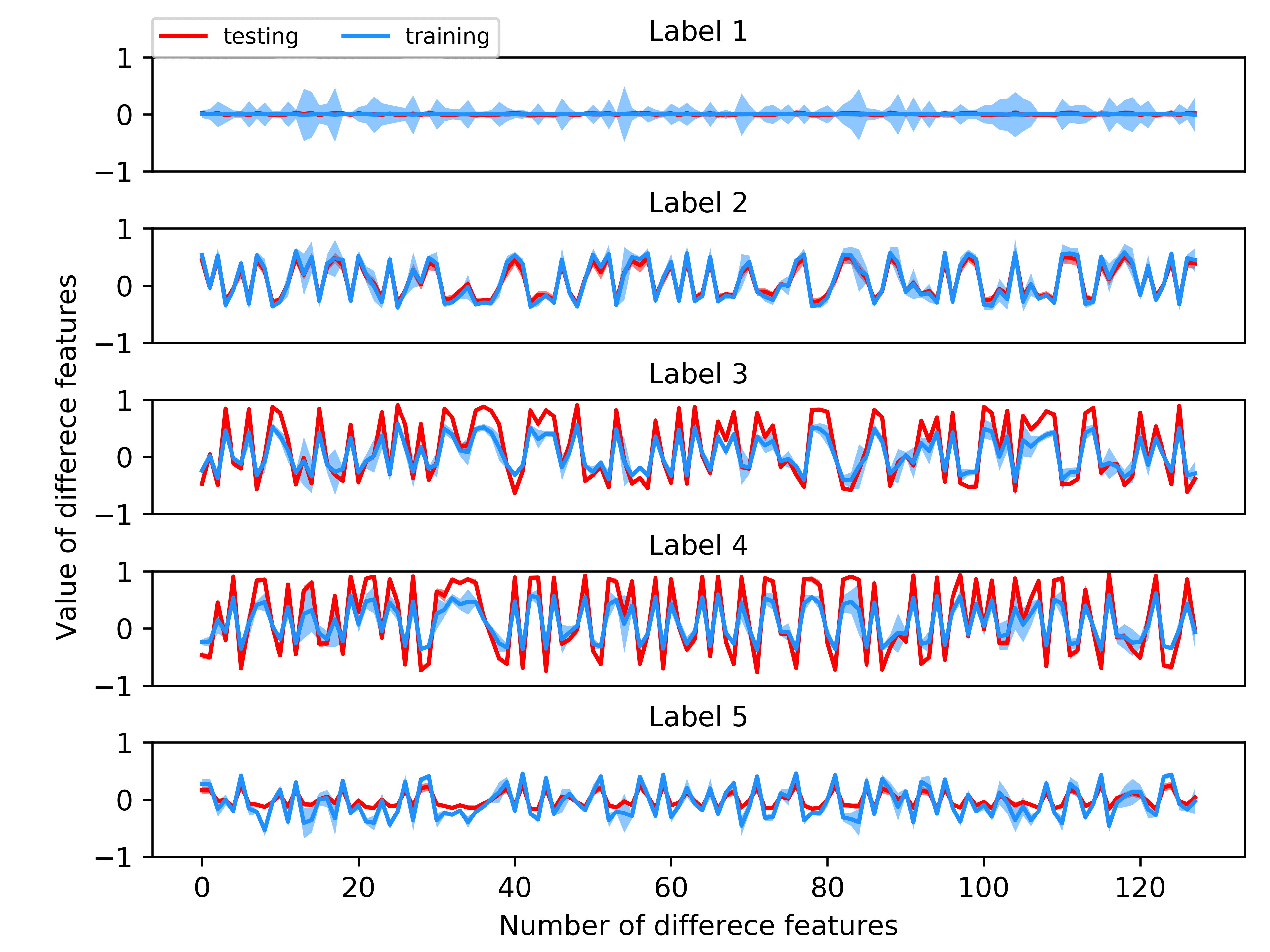}
	\caption{Visualization of difference features of each category learned by DDCNN+DA on training Dataset A (in blue) and testing Dataset C (in red).}
	\label{diff_feature}
\end{figure}

To better understand the DDCNN+DA model, the difference features  learned by DDCNN+DA on training Dataset A and testing Dataset C are plotted in Fig. \ref{diff_feature}. As shown, for each fault label, the solid line and the shaded area represents the average value and the standard deviation of the difference features calculated on 3000 training samples (in blue) and 800 testing samples (in red), respectively. It should be noted that the difference features used in the testing for the training samples are the average features of the all-healthy training samples. 

As shown, the difference features of the test samples overlap well with the corresponding features of the training samples, which explains why the DDCNN+DA model achieves high accuracy in the testing dataset. In addition, the training and testing features of the all-healthy category (Label 1) are all around zero, which is consistent with (\ref{d_test}) and (\ref{phi_average}). Moreover, as the damaged degrees of the broken Propellers of category 3 and 4 are heavier than the counterparts of category 2 and 5, the feature differences between the training and testing samples of category 3 and 4 are greater than the counterparts of category 2 and 5.

\begin{figure}[htp]
	\centering
	\includegraphics[width=0.8\linewidth]{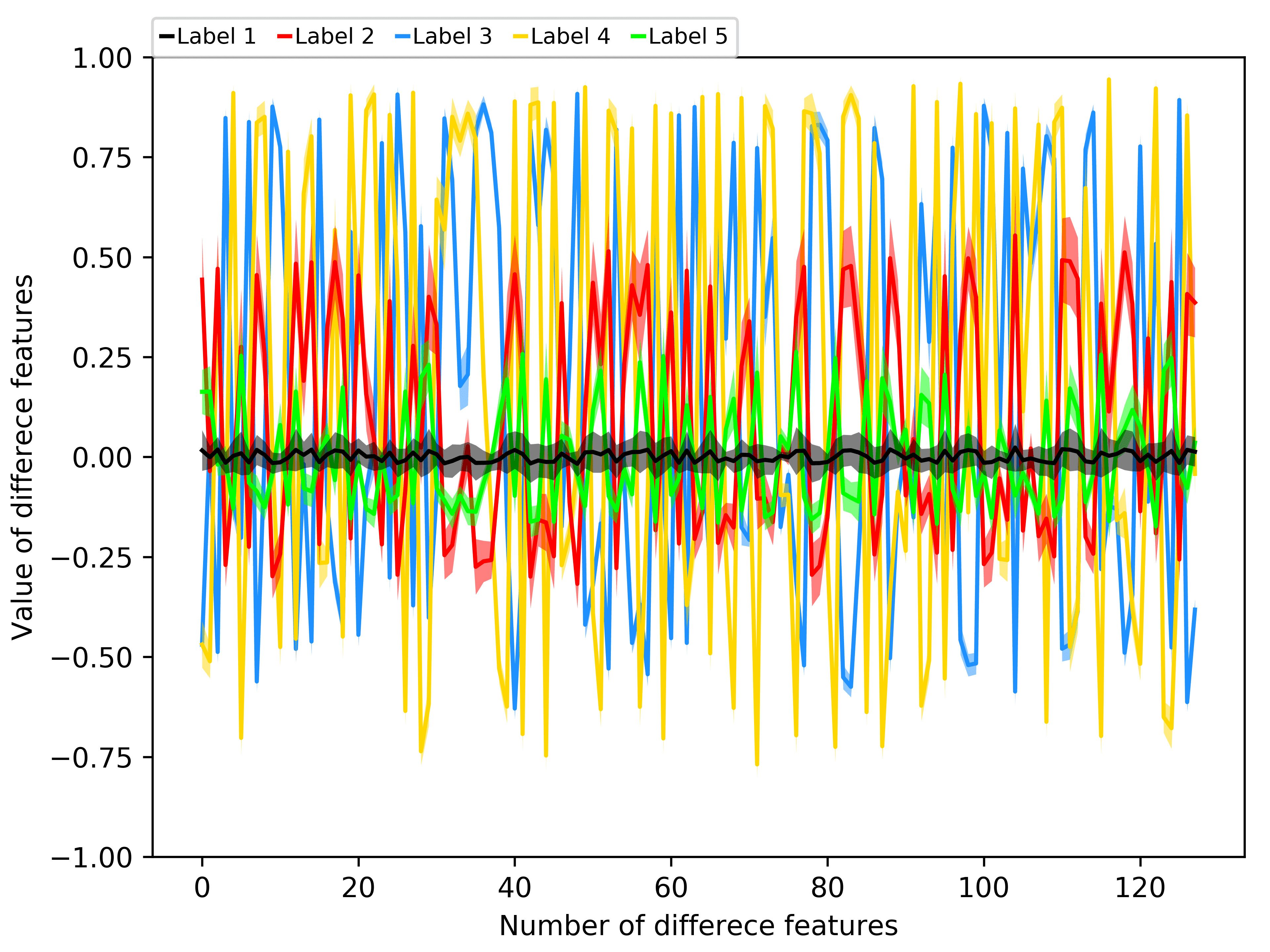}
	\caption{Visualization of all-category difference features learned by DDCNN+DA on testing Dataset C.}
	\label{diff_feature_all}
\end{figure}

In addition, all-category difference features learned by DDCNN+DA on testing Dataset C are plotted together in Fig. \ref{diff_feature_all}. As shown, the features of five categories are different from each other, which makes it easier for the classifier to make the correct classification. Moreover, the magnitudes of the difference features are different for each category. We observe that when the magnitudes of the difference features are large, the corresponding damage level of the propeller is also severe. Specifically, according to Fig. \ref{bp}, the damage level $DL$ for the five labels is $DL_3\approx DL_4>DL_2>DL_5>DL_1$, while in Fig. \ref{diff_feature_all}, the peak value $PV$ of the difference features for the five labels are $PV_3\approx PV_4>PV_2>PV_5>PV_1$ as well. This observation provides a direction for calculating the damage level of broken propellers.


\subsection{t-SNE Visualization}\label{vis}
\begin{figure}[t]
	\centering
	  \subfloat[]{
       \includegraphics[width=0.49\linewidth]{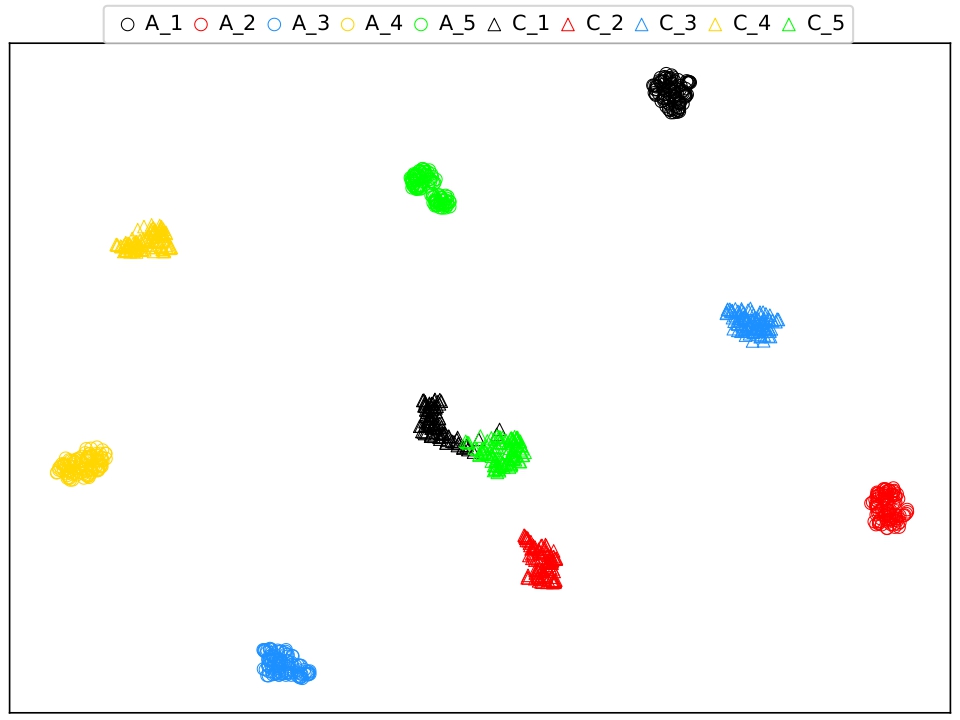}\label{tsne_DCNN}}
	  \subfloat[]{
        \includegraphics[width=0.49\linewidth]{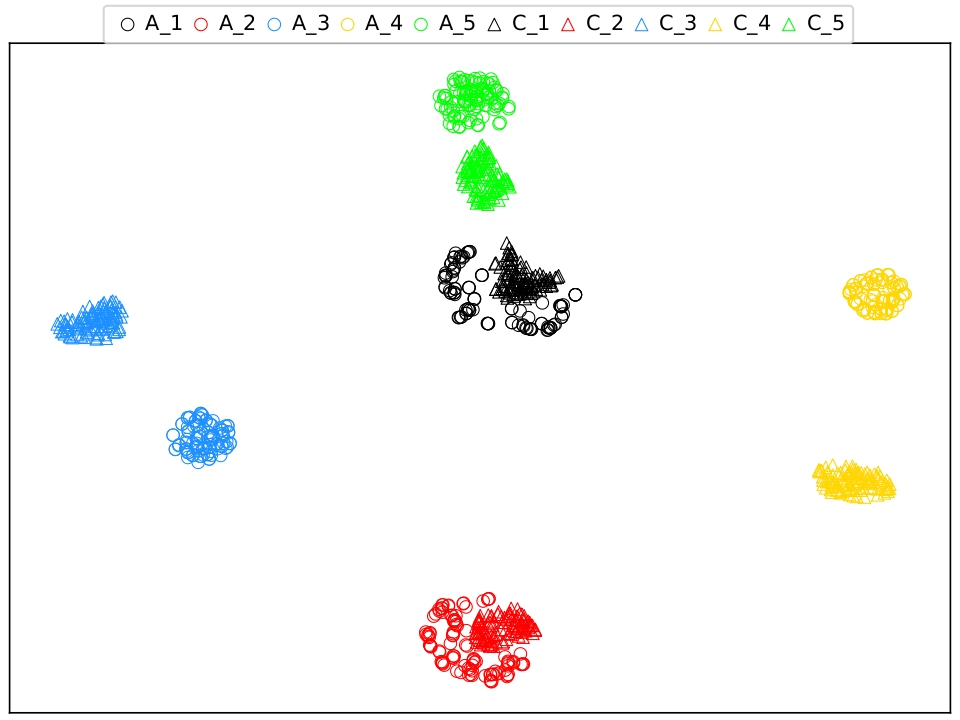}\label{tsne_DDCNN}}
	\caption{t-SNE visualization of features learned by the DCNN and DDCNN+DA model on the training datasets A and testing dataset C (the features can be better viewed via zooming in)}
	\label{real_tsne}
\end{figure}

Moreover, the difference features learned by the DCNN and DDCNN+DA model on the training datasets A and testing dataset C are compressed into 2d features via t-SNE. To better view the shape of the features, 100 samples were drawn uniformly from each fault class to generate 2d features. As shown in Fig. \ref{real_tsne}, for the DCNN model, the feature distributions of five fault categories in the target domain are far from their counterparts in the source domain. It clearly shows the domain gap and explains why the DCNN model fails to make correct decisions on real flight data. In sharp contrast, the distribution of real-flight data is closer to that of the simulation data. Due to the alignment of the two distributions, the DDCNN+DA model can achieve high accuracy on the real-flight data.

\section{Conclusion}\label{Conclusion}
In this paper, the proposed approach successfully solved the sim-to-real UAV fault diagnosis problem. With the proposed DDCNN model, the average accuracy of the classifier was greatly increased from 52.9\% to 85.5\%. With the help of the proposed DA method, the DDCNN model can achieve an average accuracy of 99.1\%. The feature visualization made it easier to understand why our DDCNN+DA model performed well in sim-to-real fault diagnosis tasks. The idea of DDCNN, which takes advantage of the difference features between faulty and healthy samples, provides a promising direction for the fault diagnosis field. In the future, we plan to employ DDCNN for diagnosing UAV faults involving multiple faulty propellers. 

\bibliographystyle{IEEEtran}
\bibliography{IEEEabrv,BIB_xx-TIE-xxxx}\ 

\end{document}